\documentclass[conference]{IEEEtran}
\IEEEoverridecommandlockouts
\usepackage{cite}
\usepackage{amsmath,amssymb,amsfonts}
\usepackage{algorithmic}
\usepackage{graphicx}
\usepackage{textcomp}
\usepackage{xcolor}
\usepackage{subfigure}
\usepackage{multirow}
\usepackage{threeparttable}
 \usepackage{booktabs}
 \usepackage{graphicx}
 \usepackage{hyperref} 
 \usepackage[letterpaper,top=54pt,bottom=54pt,left=54pt,right=54pt]{geometry}

\hypersetup{hidelinks} 
\def\BibTeX{{\rm B\kern-.05em{\sc i\kern-.025em b}\kern-.08em
    T\kern-.1667em\lower.7ex\hbox{E}\kern-.125emX}}
\begin{document}

\title{FaFCNN: A General Disease Classification Framework Based on Feature Fusion Neural Networks}

\author{
	\IEEEauthorblockN{
		Menglin Kong$^{a}$,
         Shaojie Zhao$^{b}$, 
         Juan Cheng$^{a}$, 
          Xingquan Li$^{c,d}$, 
          Ri Su$^{a}$, 
          Muzhou Hou$^{a}$, 
		Cong Cao$^{a*}$ 
\thanks{ Cong Cao is the corresponding author (congcao@csu.edu.cn).}
		}
	\IEEEauthorblockA{$^a$ School of Mathematics and Statistics,
Central South University, Changsha, China}
	\IEEEauthorblockA{$^b$ School of Mathematics, Physics \& Statistics,
Shanghai University of Engineering Science, Shanghai, China}
\IEEEauthorblockA{$^c$ Peng Cheng Laboratory, Shenzhen, China}
	\IEEEauthorblockA{$^d$ School of Mathematics and Statistics,
Minnan Normal University, Zhangzhou, China}
\IEEEauthorblockA{\{212112025,~suricsu,~hmzw,~congcao\}@csu.edu.cn,~chengjuan0306@163.com,~m440121303@sues.edu.cn,~fzulxq@gmail.com}
}

\maketitle

\begin{abstract}
There are two fundamental problems in applying deep learning/machine learning methods to disease classification tasks, one is the insufficient number and poor quality of training samples; another one is how to effectively fuse multiple source features and thus train robust classification models. To address these problems, inspired by the process of human learning knowledge, we propose the Feature-aware Fusion Correlation Neural Network (FaFCNN), which introduces a feature-aware interaction module and a feature alignment module based on domain adversarial learning. This is a general framework for disease classification, and FaFCNN improves the way existing methods obtain sample correlation features. The experimental results show that training using augmented features obtained by pre-training gradient boosting decision tree yields more performance gains than random-forest based methods. On the low-quality dataset with a large amount of missing data in our setup, FaFCNN obtains a consistently optimal performance compared to competitive baselines. In addition, extensive experiments demonstrate the robustness of the proposed method and the effectiveness of each component of the model\footnote{Accepted in IEEE SMC2023}.
\end{abstract}


\section{INTRODUCTION}
With the ability to use large amounts of precisely labelled data, deep neural network-based approaches have achieved exciting results for tasks such as e-commerce recommendation systems \cite{b1}, image classification \cite{b2}, and object detection \cite{b3}. However, many tasks in the medical field tend to have insufficient samples and a large amount of missing data, which makes it extremely difficult to develop a general deep-learning framework for disease classification tasks in the medical field \cite{b4}\cite{b5}. In addition, the records corresponding to patients in hospital databases often involve demographic features, clinical features, radiological features, and other diagnostic metrics from multiple sources; there are often scale inconsistencies and information redundancy among these data, and using them together to train machine learning models may compromise the interpretability and robustness of the models \cite{b6}. In summary, there are two fundamental problems in applying deep learning/machine learning methods to disease classification tasks: (1) the insufficient number and poor quality of training samples; (2) how to effectively fuse multiple source features and thus train robust classification models.

To address these problems, inspired by the process of human learning knowledge, some researchers \cite{b7,b8,b9} propose to augment the feature representation of each sample using the features of similar samples in the training set, i.e., introducing sample correlation features as an extension of existing features. As in \cite{b7}, the authors make an adjustment to the coupled two-stage modelling by directly using the prediction probabilities of the random forest (RF) model as correlation features with the original features as input, using a DNN with a two-tower structure to map the two parts of features separately, and finally making predictions based on the summation of high-level features. However, the prediction probability of the RF model of each sample is not enough to characterize the similarity with other samples in the training set, which will impair the performance of the model. The literature \cite{b8} proposes a graph generation method for medical datasets based on sample paths of a pre-trained random forest (RF) model, transforming structured data into graph data and training a graph convolutional network for node classification to achieve accurate differentiation of Crohn’s disease and intestinal tuberculosis. Nevertheless, this method relies heavily on artificial thresholds to determine the edges between nodes when constructing graph data, which leads to poor robustness of the framework.

AI has shown powerful potential in the field of data-driven medical fields. Esteva et al. \cite{b4} elaborated on the application prospects of various methods in the field of deep learning in the medical field from four aspects: computer vision, natural language processing, reinforcement learning and generalized deep learning methods. Rauschert et al. \cite{b10} briefly summarized the current state of Machine Learning (ML), and showed that recent advances in deep learning offer greater promise in helping physicians achieve accurate diagnoses. For example, Lima et al. proposed FSTBSVM \cite{b11}, a twin-bounded SVM classifier combined with a scalable feature selection method. And then, Kuma et al. proposed a classification algorithm that combines $k$-nearest neighbour and genetic algorithm \cite{b12}; Gu et al. proposed a fuzzy support machine with the Gaussian kernel as well as linear kernel \cite{b13}. However, due to the problems of small sample size and incomplete data in medical datasets, existing studies basically design special classification algorithms for specific disease classification tasks and basically follow the paradigm of feature selection plus machine learning model prediction. At present, there is no unified and generalized framework for auxiliary diagnosis of medical diseases.

Considering the advantages and disadvantages of existing methods, we propose the \textbf{F}eature-\textbf{a}ware \textbf{F}usion \textbf{C}orrelation \textbf{N}eural \textbf{N}etwork (FaFCNN), a general framework for disease classification. Specifically, we keep the idea of using an agent model to obtain sample correlation features to realize feature augmentation from existing methods, while the sample correlation features are acquired based on the positions of samples in the leaf nodes of a pre-trained gradient boosting decision tree (GBDT). It is experimentally demonstrated that the augmented features obtained by our method capture more accurate sample correlation than the RF-based augmented features, and further improve the model performance. In order to further improve the performance of disease classification models on low-quality datasets, FaFCNN considers the correlation of features in addition to the correlation of samples, and introduces a feature-aware interaction module (FaIM) and a feature alignment module (FAM) based on domain adversarial learning to achieve more efficient feature fusion and model performance.

The contributions of this paper are listed below:
\begin{itemize}
	\item We propose FaFCNN, a generic deep learning-based framework for disease classification, and our method obtains a consistently optimal performance compared to competitive baselines.
	\item We improve the way existing methods obtain sample correlation features, training using augmented features obtained by pre-training GBDT yields more performance gains than RF-based methods.
	\item In the feature fusion approach, the feature alignment module based on domain adversarial learning introduced by FaFCNN alleviates the performance degradation caused by the naive summation of existing methods.
	\item We synthesise low-quality datasets by adding different levels of perturbation on four public datasets. Extensive experiments demonstrate the robustness of our proposed method and the effectiveness of each component of the model.
\end{itemize}

\section{METHODOLOGY}
\subsection{Correlation Features Construction}
\begin{figure}[ht]
	\centering
	\includegraphics[width=0.42\textwidth]{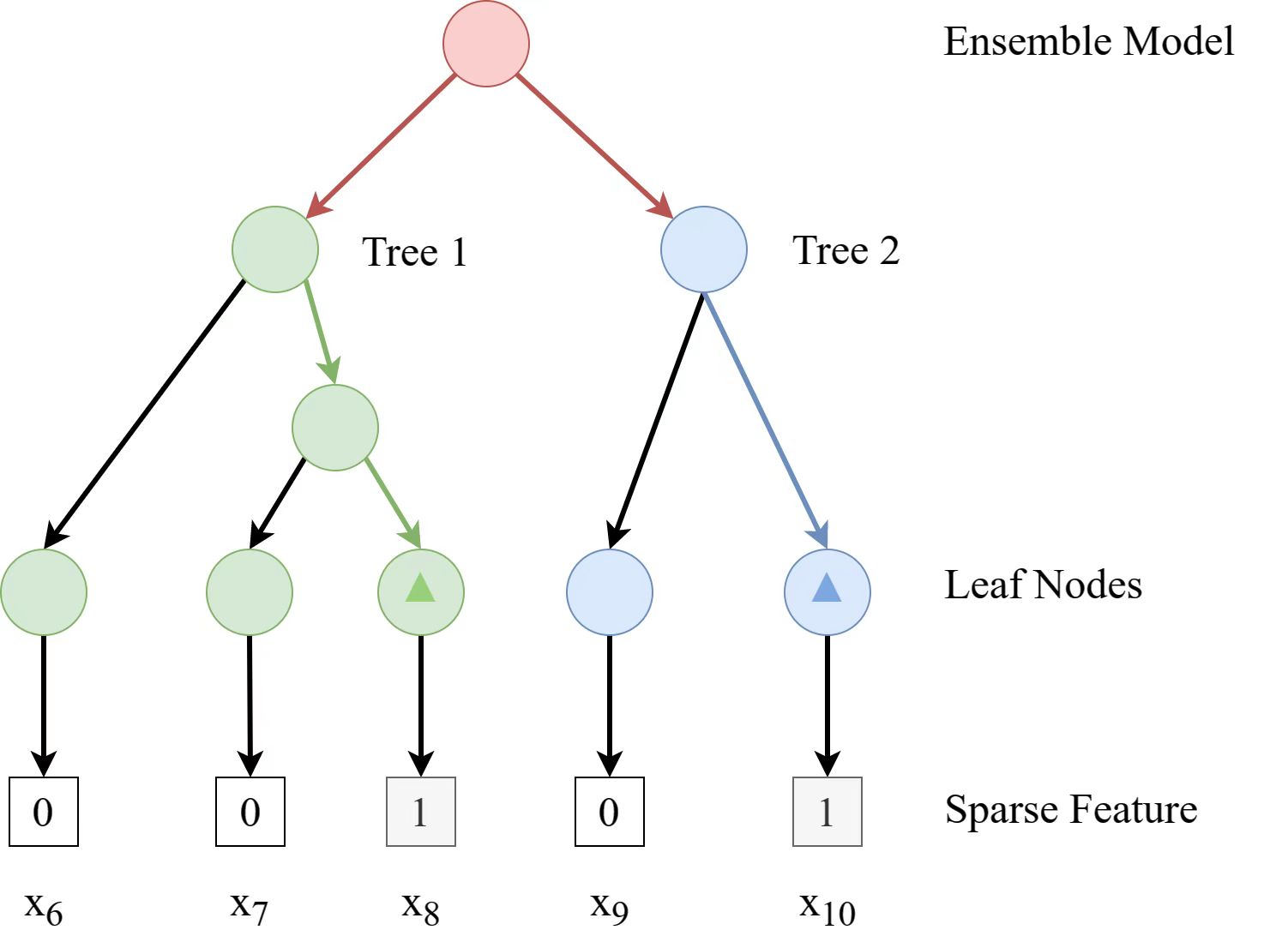}
	\caption{The diagram of correlation features construction based on GBDT. The red circle represents the root node, the green circles represents the middle node and leaf node of the first base classifier, the blue circle represents the second base classifier, and the triangles represent the position of the sample to be predicted in the leaf node of the base classifier.}\label{pic:gbdt}
\end{figure}

In this section, we present the construction of sample similarity features based on pre-trained GBDT. GBDT \cite{b14}\cite{b15} is an integrated model consisting of decision trees that learn in a gradient-boosting manner, where each base classifier (DT) is trained to fit the residuals of the prediction results of the preorder model. The GBDT structure is shown in ``Fig.~\ref{pic:gbdt}''.

In a more general scenario, the sample correlation feature construction method can be expressed as follows: first train a GBDT model on the full training data $D_{train}$ with the number of base classifiers $M$ and the number of leaf nodes of each base classifier $k$. For a sample $\mathbf{x} \in \mathbb{R}^d$ with $d$-dimensional features, it is fed into the model to get the prediction result and its position in the leaf nodes is recorded and a $k\times M$-dimensional one-hot vector is obtained, i.e., $\mathbf{x_{aug}}=(0,1,0,\cdots,0) \in \mathbb{R}^{k\times M}$, and finally the $\mathbf{x}$ is concatenated with the original features $\mathbf{x}$ to the augmented feature vector $\tilde{\mathbf{x}}\in \mathbb{R}^{k\times M+d}$. In this way, the generated one-hot vector represents the position of the sample's leaf nodes in GBDT, i.e., the predicted path of the sample given by each base classifier. From the perspective of partitioning the feature space, the prediction path of each base classifier corresponds to the subregion in the feature space where the sample points are located in a certain view. The more the intersection of the prediction paths of two samples, the more they are in the same subregion in multiple views, i.e., they have a higher correlation. We can use this correlation to augment the sample features to train a deep neural network with more powerful representation ability.

\subsection{Feature-aware Interaction Module}
FaFCNN improves the naive FCNN in two ways, respectively, by introducing FaIM to perform correlation-based mapping(i.e., feature interaction) on the sample correlation features $\mathbf{x_{aug}}$, which results in a finer-grained intermediate representation about $\mathbf{x_{aug}}$; and by introducing FAM to perform domain adversarial learning-based feature alignment operations on the original features $\mathbf{x}$ and the intermediate representation of the sample correlation features $\mathbf{x_{aug}}$.

\begin{figure*}[!ht]
	\centering
	\includegraphics[width=0.99\textwidth]{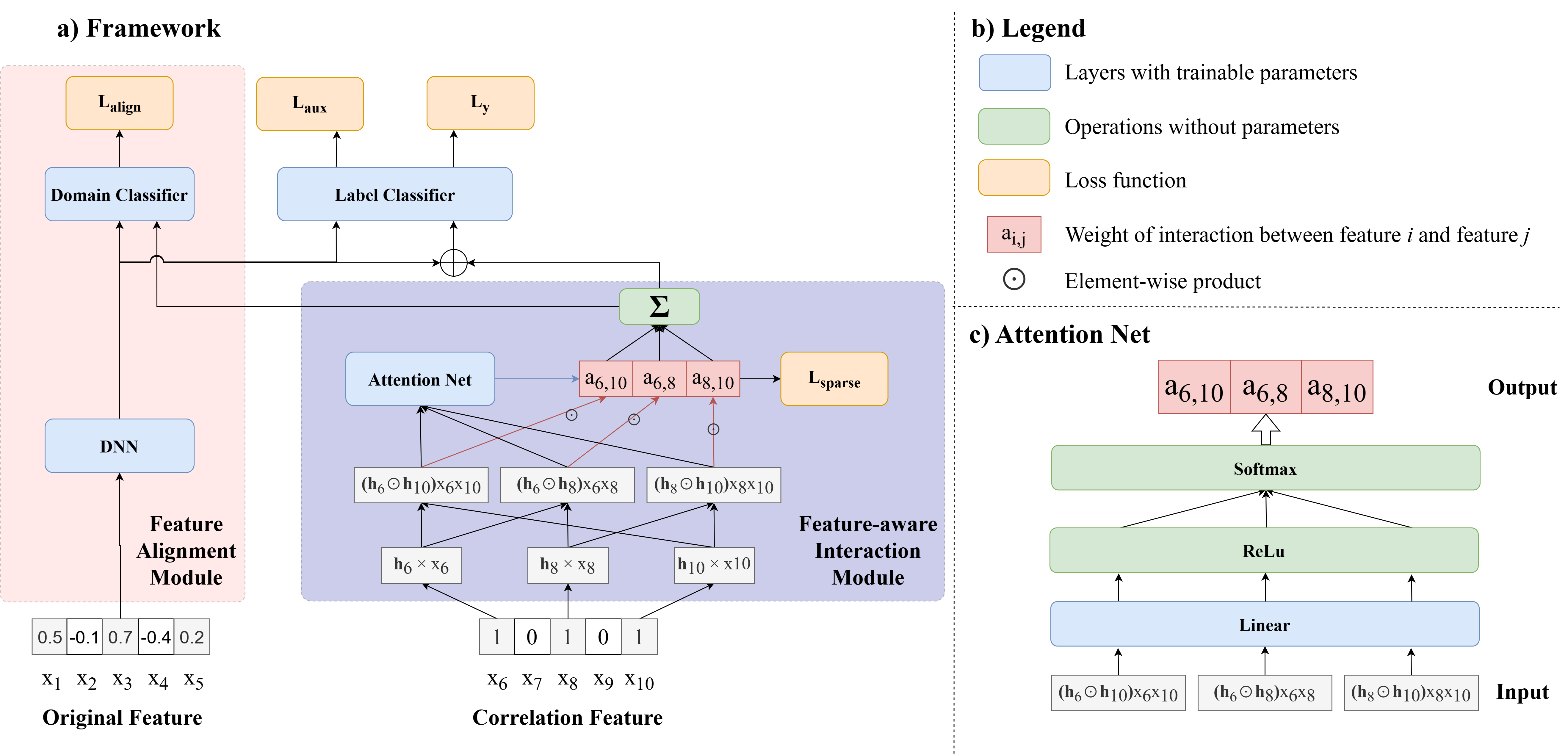}
	\caption{The sturctural diagram of the proposed FaFCNN. (a) The overall framework of FaFCNN. The part in the orange box is FAM, the part in the purple box is FaIM. (b) is an explanation of those graphics that appear previously. (c)The forward calculation process in the Attention Net of the FaIM}\label{fig:FaFCNN}
\end{figure*}

In this section, we detail how FaIM obtains fine-grained intermediate representations about sample correlation features $\mathbf{x_{aug}}$ by modeling feature interaction. As illustrated in the purple box of ``Fig.~\ref{fig:FaFCNN}'', considering a sample with 5-dimensional sample correlation feature $\mathbf{x_{aug}}=(1,0,1,0,1)$, we initialize a $p$-dimensional vector $\mathbf{h_i}$ for the $i$ th dimension in $\mathbf{x_{aug}}$ to obtain 5 $p$-dimensional vectors, each $p$-dimensional vector characterizes richer semantic information of each dimension of $\mathbf{x_{aug}}$. Then the vector $h_i$, where $i\in \left \{ i| x_i=1 \right \}$ corresponding to the non-zero position is taken to perform the second-order interactions between features in an element-wise product manner. Attention Net, a sub-network with softmax activation function, calculates the weight $a_{i,j}$ for each feature interaction term $h_i\odot h_j$ (where $(i,j) \in \left \{ m| x_m=1 \right \}$) in a self-attention manner, and finally uses this weight to aggregate these second-order interaction features to obtain the mapped sample correlation features. More generally, consider a sample of $\mathbf{x_{aug}} \in \mathbb{R}^{k\times M}$, the $p$-dimensional vector $\mathbf{h_{aug}}$ after being mapped can be obtained by the following formula:
\begin{equation}
	\mathbf{h_{aug}}=\sum_{i=1}^{k\times M} w_{i}
	x_{i}+\sum_{i=1}^{k\times M} \sum_{j=i+1}^{k\times M}
	a_{i j}\left(\mathbf{h_i} \odot \mathbf{h_j}\right) x_{i} x_{j}
	\label{eq:1}
\end{equation}
where the weight $a_{i,j}$ is calculated by the following formula:
\begin{equation}
	\begin{array}{c}
		a_{i j=}^{\prime} \mathbf{q}^{\mathrm{T}} \operatorname{Re} \operatorname{LU}\left(\varpi_{attn}\left(\mathbf{h_i} \odot \mathbf{h_j}\right) x_{i} x_{j}+b_{attn}\right), \\
		a_{i j}=\frac{\exp \left(a_{i j}^{\prime}\right)}{\sum_{(i, j) \in \mathcal{I}_{x_{aug}}} \exp \left(a_{i j}^{\prime}\right)}
	\end{array}
	\label{eq:2}
\end{equation}
where $\mathbf{q}, \varpi_{attn}, b_{attn}$ are the parameters of the sub-network, $\mathcal{I}_{\mathbf{x_{aug}}}$ denotes the set of the index in $\mathbf{x_{aug}}$.

Due to the huge number of two-by-two combinations between features, for example, for $k\times M$-dimensional features one needs to compute $C_{2}^{k\times M} = (k\times M)\times (k\times M-1)/2$ feature interaction terms and their weights, from the perspective of enhancing the interpretability of the model and reducing the computation, we want most of the interaction terms to have a weight equals to zero. This not only highlights the combination of features with the greatest impact on the prediction but also greatly reduces the computation. Inspired by the addition of L1-norm-based regularization terms to the coefficients of the linear model in LASSO regression \cite{b16}, FaFCNN adds L1-norm-based sparse regularization terms to the output $a_{i,j}$ of Attention Net in the expectation of compressing the weights of unimportant feature combinations toward the value of zero and highlighting the important ones, the formula is as follows:
\begin{equation}
	L_{sparse} = \sum_{i=1}^{k\times M} \sum_{j=i+1}^{k\times M}\left \| a_{ij} \right \|_1
	\label{eq:3}
\end{equation}

\subsection{Feature Alignment Module}
FaFCNN introduces adversarial-learning-based FAM to achieve a smoother feature fusion by aligning the distribution of mapped original features $\mathbf{x}$ and the sample correlation features $\mathbf{x_{aug}}$ in the high-dimensional representation space, which is shown in the orange box of Fig.2(a).

Similar to FCNN, a neural network with two hidden layers (DNN in Fig.2(a)) is first used to map the original features to obtain their representations in high-dimensional space $\mathbf{h} \in \mathbb{R}^p$, the formula is as follows:
\begin{equation}
	\mathbf{h}=f_{o,2}(\varpi_{o, 2}\cdot f_{o, 1}\left(\varpi_{o, 1} \cdot  \mathbf{x} +b_{o, 1}\right)+b_{o,2})
	\label{eq:4}
\end{equation}
Where $\varpi_{o, 1},\varpi_{o, 2},b_{o, 1},b_{o, 2}$ are the parameters of DNN. Since FaFCNN uses different mapping methods for different features (correlation-based aggregation for $\mathbf{x_{aug}}$, and MLP-based nonlinear mapping for $\mathbf{x}$), which leads to a large difference in the distribution of the two parts of features in the high-dimensional representation. Therefore, FaFCNN introduces the FAM module to achieve the distribution alignment of $\mathbf{h}$ and $\mathbf{h_{aug}}$ in the representation space with the idea of Min-Max game in generative adversarial networks (GAN) \cite{b17}.

FaFCNN introduces a discriminator $D$ in FAM to distinguish signals from two partial features with $\mathbf{h}$ and $\mathbf{h_{aug}}$ as inputs, and the optimization objective is to enhance the discriminator's ability to distinguish $\mathbf{h}$ and $\mathbf{h_{aug}}$, i.e:
\begin{equation}
	\theta^*=\underset{\theta} {\operatorname{argmax}} \left \| D(\mathbf{h};\theta) - D(\mathbf{h_{aug}};
	\theta) \right \|_1 \label{eq:5}
\end{equation}
which is equal to minimizing the following formula:
\begin{equation}
	L_D=- \sum_{i=1}^{N}  \left \| D(\mathbf{h_i};\theta) - D(\mathbf{h_{aug,i}};
	\theta) \right \|_1
	\label{eq:6}
\end{equation}
where $\theta$ is the parameter of the $D$, which is a two-layer MLP in FaFCNN. Naturally, we can consider the above-mentioned DNN that maps $\mathbf{x}$ as the generator $G$ in GAN, whose optimization goal is to make the distribution of the mapped $\mathbf{h}$ in the high-dimensional representation space as similar as possible to $\mathbf{h_{aug}}$, so that the discriminator $D$ cannot distinguish $\mathbf{h_{aug}}$ from $\mathbf{h}$, the formula is as follows:

\begin{equation}
	\phi^*=\underset{\phi} {\operatorname{argmin}}
	\left \| D(G(\mathbf{x};\phi);\theta) - \mathbf{1}\right \|_1
	\label{eq:7}
\end{equation}
where $\phi=\left \{ \varpi_{o, 1},\varpi_{o, 2},b_{o, 1},b_{o, 2} \right \} $ is the parameter of the DNN, $\theta^*$ is the optimal parameters of the discriminator in the last iteration, $\mathbf{1} \in \mathbb{R}^p$ is an all-one vector(proxy label of $\mathbf{h_{aug}}$ in this domain adversarial learning procedure). This optimal $\phi^*$ can be found by minimizing the following loss function:
\begin{equation}
	L_G=\sum_{i=1}^{N}  \left \| D(G(\mathbf{x_i};\phi);\theta) - \mathbf{1}\right \|_1
	\label{eq:8}
\end{equation}
However, it is easy to experience pattern collapse during training with GAN, i.e., the generator generates very narrow distributions that cover only a single pattern in the data distribution. This was also observed in our experiments, where DNNs tend to consistently map samples with different original features $\mathbf{x}$ to limited range in the high-dimensional space during adversarial learning, but since this is a pattern in the $\mathbf{h_{aug}}$ distribution, the purpose of tricking the discriminator $D$ can be reached. While this single-pattern representation does not bring any beneficial information to the model for classification. To ensure that the aligned $\mathbf{h}$ maintains diversity during the adversarial learning, FaFCNN introduces a supervised signal to $\mathbf{h}$ so that it retains a certain amount of information that is beneficial to the classification of the sample thus ensuring that the aligned $\mathbf{h}$ diversity of the distribution pattern, noting the label classifier as $F(\cdot;\psi)$ and the auxiliary loss as follows:
\begin{equation}
	\begin{aligned}
		L_{aux}=-\frac{1}{N} \sum_{i=1}^{N} ( & y_{i} \log F(\mathbf{h_i};\psi) + \\
		& \left ( 1-y_{{i}} \right ) \log \left (1-F(\mathbf{h_i};\psi) \right ) )
	\end{aligned}
	\label{eq:9}
\end{equation}

\subsection{Optimization}
Based on the above, the training process of FaFCNN consists of two stages. In the first stage, we train the $\mathbf{h_{aug}}$ obtained from FaIM to make it capable of classifying samples in a supervised manner, while adding a sparse regularization term to the total loss to ensure the sparsity of the weights of feature interaction terms obtained from Attention Net, the formula is as follows:
\begin{equation}
	\begin{aligned}
		L_{y}=-\frac{1}{N} \sum_{i=1}^{N} (& y_{i} \log F(\mathbf{h_{i,aug}};\psi) + \\
		& \left (1-y_{{i}} \right ) \log \left ( 1-F(\mathbf{h_{i,aug}};\psi) \right ) )
	\end{aligned}
	\label{eq:14}
\end{equation}
\begin{equation}
	L_{1}=L_y+\alpha L_{sparse}  \label{eq:15}
\end{equation}
In the second stage, we first freeze the network parameters in the already trained FaIM module to ensure that $\mathbf{h_{aug}}$ does not change during FAM training. Then alternately optimize the parameters $\theta$ of the discriminator $D$ with Equation \ref{eq:6}, and the parameters $\phi$ of the DNN with Equation \eqref{eq:8} and \eqref{eq:9}, the formula is as follows:
\begin{equation}
	L_2 = L_{aux}+\beta L_G
	\label{eq:16}
\end{equation}

\section{EXPERIMENTS}
In this section, We validate the effectiveness and robustness of FaFCNN on four publicly available medical datasets with special perturbation treatments.

\subsection{Experimental Setting.}
\subsubsection{Dataset}
To prove the superiority of the proposed method in medical diagnosis, we apply our model to four public medical datasets, including the Wisconsin Breast cancer, Pima Indians Diabetes, Hepatitis, Heart-Statlog datasets, more details of these datasets are listed as follows:

\begin{table}[ht]
	\centering
	\small
	\fontsize{9}{12}\selectfont
	\caption{Public datasets description}\label{table:1}
	\resizebox{0.48\textwidth}{!}{%
		\begin{tabular}{@{}ccccc@{}}
			\toprule
			\textbf{Dataset}        & \textbf{N-smaples} & \textbf{N-features} & \textbf{N-classes} & \textbf{Reference} \\ \midrule
			Wisconsin Breast Cancer & 699                & 9                   & 2                  & \cite{b18}            \\
			Pima Indians Diabetes   & 768                & 8                   & 2                  & \cite{b19}            \\
			Hepatitis               & 155                & 19                  & 2                  & \cite{b20}            \\
			Heart-Statlog           & 270                & 13                  & 2                  & \cite{b21}            \\ \bottomrule
		\end{tabular}%
	}
\end{table}
To simulate the challenge of existing a large number of missing values in a real scenario medical dataset, we add different levels of perturbation to the above dataset. Considering a raw dataset with $N$ samples and $d$ features, the data preprocessing process is as follows:
\begin{itemize}
	\item First, the missing values in the dataset are processed. The columns with missing values are first identified and the median of the column other than the missing values is calculated and the missing values are replaced by the median.
	\item Then, the dataset is perturbed randomly. The data are first shuffled by rows, and then the rows of data to be perturbed are removed according to the selected $\delta$. Each column of these data rows is randomly selected with equal probability ($1/d$) and perturbed in the same way as the missing values are processed.
	\item Finally, the data set is divided into a training set, validation set, and test set in the ratio of 8:1:1.
\end{itemize}

\subsubsection{Hyperparameter}
In this section the hyperparameter settings used for training FaFCNN are described, the parameters of the GBDT included $k=integer(d/2)$($d$ is the number of features of the dataset) estimators, 8 max depths, and $M=2$ min sample leaves. The stage 1 training of FaFCNN uses the Adam optimizer with a learning rate of 0.005 for $T_1 = 10000$ epochs. The stage 2 training of DNN and Discriminator uses the SGD optimizer with a learning rate of 0.005 for $T_2 = 10000$ epochs. The dimension $p$ of vector $\mathbf{h}$ is set as 8, and the balance coefficient $\alpha$ and $\beta$ are set to 0.05 and 0.5, respectively.

\subsubsection{Performance evaluation metrics}
We select accuracy, sensitivity and Specificity, which are common evaluation metrics in classification tasks, to compare the performance of FaFCNN and baseline models in three dimensions. The three evaluation indicators are defined as follows:

{\setlength\abovedisplayskip{1pt}
\setlength\belowdisplayskip{1pt}
\begin{subequations}
\begin{align}
&Acc = \frac{\mathcal{M}_{tp} + \mathcal{M}_{tn}}{\mathcal{M}_{tp} + \mathcal{M}_{fp} + \mathcal{M}_{tn} + \mathcal{M}_{fn}}\\
&Sensitivity = \frac{\mathcal{M}_{tp}}{\mathcal{M}_{tp} + \mathcal{M}_{fn}}\\
&Specificity = \frac{\mathcal{M}_{tn}}{\mathcal{M}_{tn} + \mathcal{M}_{fp}}
\end{align}
\label{eq:metrics}
\end{subequations}}

\subsection{Classification results of different classification methods}
Our comparison is performed uniformly on four datasets with a perturbation ratio of $\delta=0.5$; to ensure the fairness of the comparison, the structure of the DL-based baseline is adjusted so that the number of parameters of the models involved in the comparison remains the same. The experimental results are shown in Table \ref{table:2}, and results in the table are the mean values of 10 independent repetitions of the experiment.

\begin{table*}[ht]
	\centering
	\small
	\fontsize{9}{12}\selectfont
	\caption{Performance results of different models on large competition data sets}\label{table:2}
	\resizebox{\textwidth}{!}{%
		\begin{tabular}{|cc|cccccccccccc|}
			\hline
			\multicolumn{2}{|c|}{\multirow{3}{*}{\textbf{Methods}}}                                    & \multicolumn{12}{c|}{\textbf{Datasets}}                                                                                                                                                                                                                                    \\ \cline{3-14}
			 \multicolumn{2}{|c|}{}                  & \multicolumn{3}{c|}{Wisconsin Breast Cancer}                            & \multicolumn{3}{c|}{Pima Indians Diabetes}                              & \multicolumn{3}{c|}{Hepatitis}                                         & \multicolumn{3}{c|}{Heart-Statlog}                 \\ \cline{3-14}
		\multicolumn{2}{|c|}{}                                                      & Accuracy(\%)   & Sensitivity(\%) & \multicolumn{1}{c|}{Specificity(\%)} & Accuracy(\%)   & Sensitivity(\%) & \multicolumn{1}{c|}{Specificity(\%)} & Accuracy(\%)  & Sensitivity(\%) & \multicolumn{1}{c|}{Specificity(\%)} & Accuracy(\%)   & Sensitivity(\%) & Specificity(\%) \\ \hline
			\multicolumn{1}{|c|}{\multirow{7}{*}{\textbf{ML-based}}} & LR                            & 90.7           & 79.5            & \multicolumn{1}{c|}{86.1}            & 76.6           & 51.8            & \multicolumn{1}{c|}{76.3}            & 83.5          & 67.2            & \multicolumn{1}{c|}{92.5}            & 87             & 71.4            & 93.8            \\
			\multicolumn{1}{|c|}{}                                   & RF                            & 92.1           & 91.5            & \multicolumn{1}{c|}{86}              & 76.6           & 49              & \multicolumn{1}{c|}{68.6}            & 86.7          & 85.5            & \multicolumn{1}{c|}{71}              & 83.3           & 71.4            & 83.3            \\
			\multicolumn{1}{|c|}{}                                   & PSO-ELM\cite{b8}       & 93.6           & 89.8            & \multicolumn{1}{c|}{91.7}            & 78.6           & 56.4            & \multicolumn{1}{c|}{77.5}            & 97.4          & 93.7            & \multicolumn{1}{c|}{95.7}            & 85.9           & 86              & 86              \\
			\multicolumn{1}{|c|}{}                                   & SRLPSO-ELM\cite{b8}    & 91.4           & 84.6            & \multicolumn{1}{c|}{84.6}            & 74             & 50              & \multicolumn{1}{c|}{65}              & \textbf{98.7} & 94.2            & \multicolumn{1}{c|}{96}              & 89.9           & 87.8            & 88.4            \\
			\multicolumn{1}{|c|}{}                                   & FSTBSVM\cite{b11}       & 95             & 92.3            & \multicolumn{1}{c|}{90}              & 76.6           & 69.2            & \multicolumn{1}{c|}{64.3}            & 89.1          & 78.6            & \multicolumn{1}{c|}{\textbf{100}}    & 85.9           & 72.7            & 89.8            \\
			\multicolumn{1}{|c|}{}                                   & KNN-GA\cite{b12}        & 90             & 84.8            & \multicolumn{1}{c|}{84.8}            & 72.1           & 33.9            & \multicolumn{1}{c|}{83.3}            & 64.4          & 78.3            & \multicolumn{1}{c|}{50}              & 83.5           & 87.5            & 82.3            \\
			\multicolumn{1}{|c|}{}                                   & FSVM-Gaussian\cite{b13} & 92.9           & 90.7            & \multicolumn{1}{c|}{86.7}            & 76             & 52.2            & \multicolumn{1}{c|}{61.5}            & 84            & 89.6            & \multicolumn{1}{c|}{87.4}            & 84.7           & 86              & 83.1            \\ \hline
			\multicolumn{1}{|c|}{\multirow{4}{*}{\textbf{DL-based}}} & RFG-GCN\cite{b8}              & 90             & 80.9            & \multicolumn{1}{c|}{88.4}            & 83.5           & 67.2            & \multicolumn{1}{c|}{93.5}            & 95.9          & {\underline{98.1}}      & \multicolumn{1}{c|}{92.2}            & 92.1           & 90.2            & 93.5            \\
			\multicolumn{1}{|c|}{}                                   & DNN\cite{b7}                  & 90             & 85.1            & \multicolumn{1}{c|}{85.1}            & 84.4           & 69.2            & \multicolumn{1}{c|}{93.4}            & 91            & 93.8            & \multicolumn{1}{c|}{89.1}            & 92.6           & 89.7            & {\underline{94.8}}      \\
			\multicolumn{1}{|c|}{}                                   & FCNN\cite{b7}                 & {\underline{95.7}}     & {\underline{97.5}}      & \multicolumn{1}{c|}{\underline{88.6}}      & {\underline{88.3}}     & {\underline{78.1}}      & \multicolumn{1}{c|}{\underline{93.9}}      & 90.8          & 95.7            & \multicolumn{1}{c|}{87.3}            & {\underline{95.1}}     & {\underline{91.3}}      & 93.5            \\
			\multicolumn{1}{|c|}{}                                   & \textbf{FaFCNN(ours)}                  & \textbf{97.9*} & \textbf{97.9}   & \multicolumn{1}{c|}{\textbf{95.8*}}  & \textbf{93.1*} & \textbf{91.5**} & \multicolumn{1}{c|}{\textbf{96.3*}}  & {\underline{98.6}}    & \textbf{98.7}   & \multicolumn{1}{c|}{\underline{98.5}}      & \textbf{97.7*} & \textbf{94.9*}  & \textbf{95.2}   \\ \hline
		\end{tabular}%
	}
\begin{tablenotes}
	\footnotesize
	\item Accuracy comparison of ours and baseline models on four 50\% perturbed datasets. The best is in bold font,  * and ** indicate $p<0.005$ and $p<0.001$ for a one-tailed t-test, and the underlining indicates sub-optimal performance. The results are averaged over 10 independent repeats on all the datasets.
\end{tablenotes}
\end{table*}

As shown in Table \ref{table:2}, in the comparison of results from the Wisconsin Breast Cancer dataset, the DL-based methods (DNN, RFG-GCN) do not show better performance in some metrics than the ML-based methods (RF, LR); the FCNN better exploits the sample correlation in the training set to achieve a consistent performance improvement over the DNN. FaFCNN achieves smoother and more effective feature fusion while considering feature correlation and achieves significant improvement in two evaluation metrics compared to FCNN.

ML-based methods perform poorly on the Pima Indians Diabetes dataset, especially the sensitivity metric does not exceed 70\% at the highest; DL-based methods achieve improvements in the other two metrics, but the sensitivity metric still can not exceed 80\% (78.1\% for FCNN). This indicates that there is a serious class imbalance problem on the Diabetes dataset and the model easily misclassifies some of the positive cases as negative cases. FaFCNN significantly outperforms FCNN with a p-value of 0.001 at 91.5\% on the sensitivity metric while the other two metrics significantly outperform the optimal baseline with a p-value of 0.005.

On the Hepatitis dataset, the SRLPSO-ELM method achieves optimal performance in accuracy but this advantage was not significant (98.7\%) and our FaFCNN has achieved (98.6\%); FSTBSVM reaches 100\% in specificity but does not significantly outperform FaFCNN (98.5\%), and this advantage comes at the expense of sensitivity (78.6\%), while FaFCNN achieves optimal performance in this metric (98.7\%).

On the Heart-Statlog dataset, DL-based methods consistently outperform ML-based methods, and FaFCNN again achieves optimal performance on the three metrics with the same number of parameters and demonstrates significance in terms of accuracy and sensitivity due to the well-designed structure of the network. In summary, our FaFCNN is able to show robust and consistent optimal performance with respect to the baseline models on multiple datasets with 50\% of the samples perturbed in a low-quality data setting and with class imbalance problems.

\subsection{Robustness verification of classification results}
\begin{figure}[!h]
	\centering
	\includegraphics[width=0.42\textwidth]{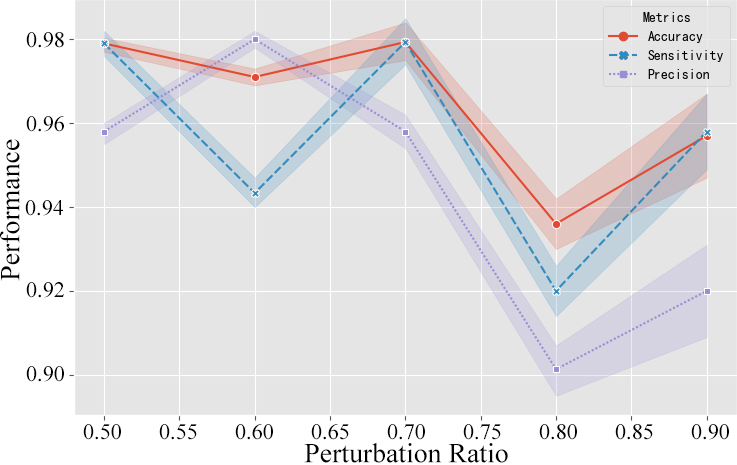}
	\caption{Performance of FaFCNN on the Wisconsin Breast Cancer dataset with different settings of perturbation ratio $\delta$. The red solid line, blue dashed line, and purple dotted line represent accuracy, sensitivity, and precision, respectively, and the points on the axes represent the mean values of 10 experiments, while the upper and lower bandwidths represent the standard deviation of the experimental results.}\label{fig:3}
\end{figure}

To verify that our proposed FaFCNN maintains robustness and acceptable performance in the face of scenarios with large amounts of missing data, we set a set of perturbation ratios $\delta \in \left \{ 0.5,0.6,0.7,0.8,0.9 \right \}$, 10 experiments are conducted on the Wisconsin Breast Cancer dataset for each $\delta$, and the results of the experiments are shown in ``Fig.~\ref{fig:3}''.

As shown in ``Fig.~\ref{fig:3}'', we can conclude that as $\delta$ gradually increases, meaning that the proportion of samples with missing values in the dataset increases, the fluctuation of FaFCNN's performance also gradually increases (the bandwidth on both sides of the performance line increases), but the three evaluation metrics still maintain a high level (the mean value of the worst case also remains above 0.9). Specifically, accuracy does not decrease significantly as $\delta$ increases, and the mean value of each case remains above 0.93. Sensitivity and Precision show a decreasing trend (when $\delta$ increases from 0.7 to 0.8) and show a performance increase when $\delta$ improves to 0.9. In summary, FaFCNN has a narrow bandwidth on both sides of the line for different values of $\delta$, which proves the robustness of the classification results in each setting; its classification performance does not show an obviously decreasing trend with the increase of $\delta$, which proves that the model has strong robustness to noisy data.

\subsection{Ablation Study} 
In this section, we focus on validating the effectiveness of the well-designed components in FaFCNN by means of ablation experiments; to ensure fairness of the comparison, each variant of FaFCNN is extended in terms of network structure to ensure consistent overall model parameters. We conduct 10 independent repetition experiments on the Wisconsin Breast Cancer dataset under the setting of $\delta=0.5$.

\subsubsection{Validity of FaIM\&FAM} 
\begin{figure}[!h]
	\centering
	\includegraphics[width=0.35\textwidth]{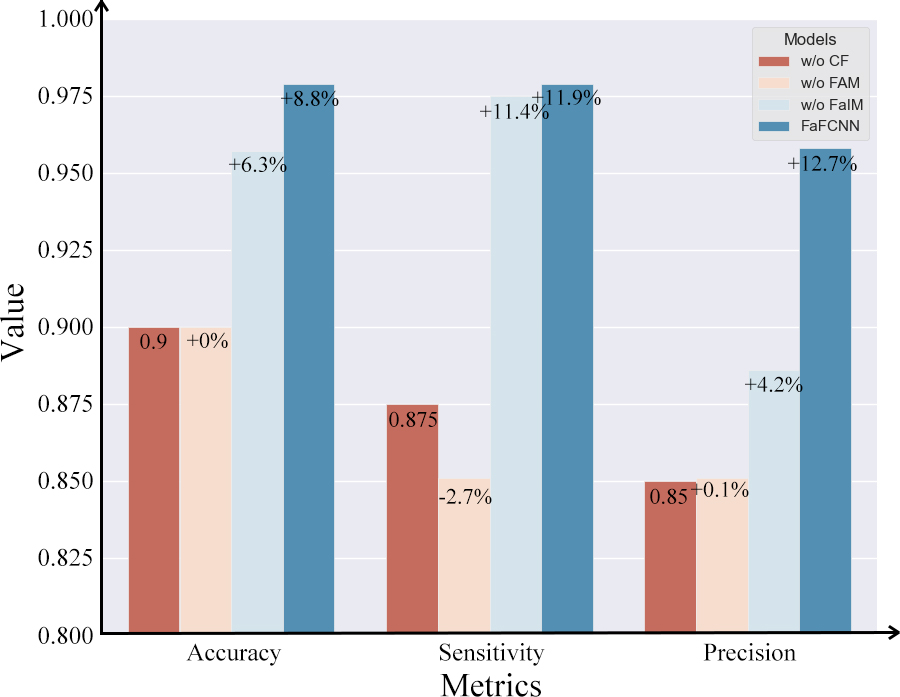}
	\caption{Comparative results of FaFCNN and its three variants on Wisconsin Breast Cancer. Dark red means no sample correlation features are used, light orange means sample correlation features are added but FAM is not used, light blue means correlation between features is not modelled using FaIM, and dark blue means FaFCNN. The number above the bar indicates the relative improvement of adding different modules compared to the base model without introducing sample correlation.}\label{fig:4}
\end{figure}

We first validate the effectiveness of the proposed modules and quantify the performance improvement brought by each module through a set of comparison experiments between FaFCNN and its three variants on the Wisconsin Breast Cancer dataset, as shown in ``Fig.~\ref{fig:4}''.

 As shown in ``Fig.~\ref{fig:4}'', compared with the base model without sample correlation features, using the output of pre-trained RF as augmented features do not improve the performance of the model, but decreases the sensitivity of the model (-2.74\%), which indicates that the introduction of augmented features without using a reasonable feature fusion method will harm the performance of the model. The introduction of FAM significantly improves the performance of the model, with improvements of 6.3\%, 11.4\%, and 4.2\% for the three metrics respectively, which validates the effectiveness of our proposed feature fusion module based on adversarial learning. FaFCNN uses the FaIM module to replace the w/o FaIM variant's DNN for mapping sample correlation features and achieves further improvements in accuracy(+8.8\%), sensitivity(+11.9\%) and precision(+12.7\%) metrics, which verifies that using predicted paths of samples of GBDT as augmented features can capture more accurate sample correlation than the RF-based approach, and the feature-interaction-based explicit mapping approach can achieve finer-grained feature representation than the DNN-based implicit feature mapping.

\subsubsection{Effectiveness of Sparse Regularization} 
\begin{figure}[!h]
	\setlength{}
	\centering
	\addtocounter{subfigure}{0}\subfigure[Without sparse regularization]{
		\includegraphics[width=.2\textwidth]{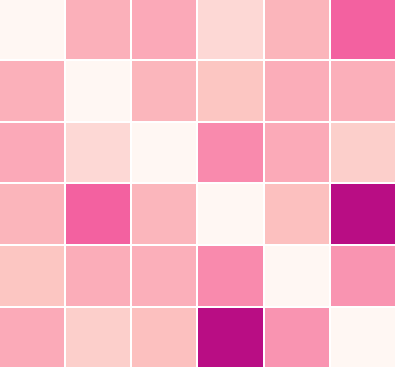}
	}
	\addtocounter{subfigure}{0}\subfigure[With sparse regularization]{
		\includegraphics[width=.2\textwidth]{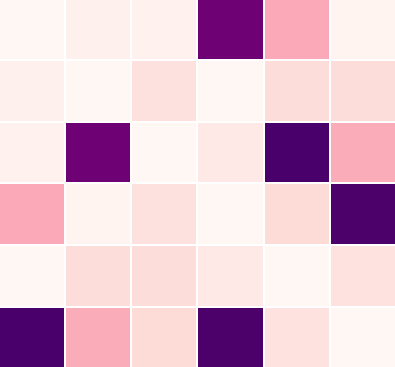}
	}
	\caption{The heat map of average weights of feature interactions in FaIM, calculated in the test phase of 50\% perturbed Wisconsin Breast Cancer dataset. The darker the color, the greater the absolute value of the weight.}
	\label{fig:5}
\end{figure}

To verify the effectiveness of the weight sparse regularization term added to the FaIM module, we train FaFCNN and FaFCNN without sparse regularization on the 50\% perturbed Wisconsin Breast Cancer dataset, respectively, and record the output of the Attention Network, i.e., the weights of the feature interactions for each sample in the test phase, then average on them. ``Fig.~\ref{fig:5}'' shows the heatmap based on the mean value of 10 repetitions of the above procedure.

FaFCNN and FaFCNN-w/o sparse regularization perform consistently on 10 independent replicate experiments, that using  sparse regularization with mean values of 97.9\%, 97.9\%, 95.8\% and Without sparse regularization with mean values of 95\%, 92\%, 93.9\% for the three metrics, respectively, and FaFCNN shows a significant improvement in accuracy and sensitivity relative to the variant without sparse regularization term. On the other hand, the above heat map shows that the two models capture similar feature association patterns, such as the larger values of $a_{3,5}$, which implies that the feature interaction terms at these two locations have a greater impact on the model prediction thus the two features are more correlated. In addition, the sparse regularization term does work as expected by reducing the weights of relatively unimportant feature interactions while increasing the weights of critical feature interactions (Fig.5(b) has many more blank squares than Fig.5(a) but with darker colors at important positions), allowing the model to discover significant feature interaction patterns in the data, thus reducing the computation by using only the important feature combinations in the subsequent modelling process.

\section{Conclusions}
In this work, by considering the advantages and disadvantages of existing methods, we propose the FaFCNN, a general framework for disease classification. On the one hand, FaFCNN improves the way existing methods obtain sample correlation features, exploiting augmented features obtained by pre-training gradient boosting decision trees to capture more accurate correlations between samples in the training set. On the other hand, FaFCNN introduces a feature alignment module for smoother and more efficient feature fusion, and the feature-aware interaction module considers feature correlation and model feature interaction in a more fine-grained manner to enhance the model's representation ability. Extensive experimental results show that FaFCNN has strong robustness and can achieve consistent optimal performance concerning the baseline models on multiple datasets with 50\% of the samples perturbed in a low-quality data setting and with class imbalance problems.

\textbf{Acknowledgements} This study was supported by Natural Science Foundation of Hunan Province of China(grant number 2022JJ30673) and by the Graduate Innovation Project of Central South University (2023XQLH032, 2023ZZTS0304).


\end{document}